\documentclass{article}
\usepackage{ijcai25}
\usepackage{natbib}

\usepackage{times}
\usepackage{soul}
\usepackage{url}
\usepackage[hidelinks]{hyperref}
\usepackage[utf8]{inputenc}
\usepackage[small]{caption}
\usepackage{graphicx}
\usepackage{amsmath}
\usepackage{amssymb}  

\usepackage{amsthm}
\usepackage{booktabs}
\usepackage{algorithm}
\usepackage{algorithmic}
\usepackage[switch]{lineno}
\usepackage{xcolor}
\usepackage{subcaption}

\title{CENTS: Generating synthetic electricity consumption time series for rare and unseen scenarios
}

\author{
    Michael Fuest$^{1,3}$\thanks{Work done as a Visiting Student at MIT.},
    Alfredo Cuesta$^2$,
    Kalyan Veeramachaneni$^3$\\
    \affiliations
    $^1$ Technical University of Munich\\
    $^2$ Universidad Rey Juan Carlos\\
    $^3$ Massachusetts Institute of Technology\\
    \emails
    michael.fuest@tum.de, alfredo.cuesta@urjc.es, kalyan@csail.mit.edu
}

\begin{document}

\maketitle

\begin{abstract}
Recent breakthroughs in large-scale generative modeling have demonstrated the potential of foundation models in domains such as natural language, computer vision, and protein structure prediction. However, their application in the energy and smart grid sector remains limited due to the scarcity and heterogeneity of high-quality data. In this work, we propose a method for creating high-fidelity electricity consumption time series data for rare and unseen context variables (e.g. location, building type, photovoltaics). Our approach, \textbf{C}ontext \textbf{E}ncoding and \textbf{N}ormalizing \textbf{T}ime \textbf{S}eries Generation, or \textbf{CENTS}, includes three key innovations: (i) A context normalization approach that enables inverse transformation for time series context variables unseen during training, (ii) a novel context encoder to condition any state-of-the-art time-series generator on arbitrary numbers and combinations of context variables, (iii) a framework for training this context encoder jointly with a time-series generator using an auxiliary context classification loss designed to increase expressivity of context embeddings and improve model performance. We further provide a comprehensive overview of different evaluation metrics for generative time series models. Our results highlight the efficacy of the proposed method in generating realistic household-level electricity consumption data, paving the way for training larger foundation models in the energy domain on synthetic as well as real-world data.
\end{abstract}

\section{Introduction}
\label{sec:introduction}

Recent advances in generative deep learning models have sparked the creation of so-called foundation models, large-scale models trained on extensive datasets that are adaptable to many downstream tasks across many domains \citep{bommasani2021opportunities}. These models have been successfully applied in natural language processing \citep{achiam2023gpt}, computer vision \citep{rombach2022high}, and protein 3D structure prediction \citep{jumper2021highly}. A common feature among foundation models is their reliance on comprehensive, high quality datasets. In the energy and smart grid domain, high-quality datasets are notoriously rare, hindering the development of foundation models in the space. In this work, we focus on addressing data shortage. Our goal is to design a framework that lets users sample generative models to produce high-quality household-level electricity consumption (and generation) data for a wide range of scenarios. Scenarios here refer to diverse combinations of household characteristics, which we refer to as context variables. We focus on household-level data, since households represent the lowest level of energy consumption modeling and can be aggregated to model consumption patterns of area codes or even larger geographies. Context variables may include location, building type or photovoltaics, but their selection is determined by the availability of household metadata in electricity consumption datasets. \emph{Rare scenarios} arise when specific combinations of context variables, such as a particular building type in a particular location, appear only infrequently in the dataset and produce consumption time series patterns that diverge substantially from the rest of the data. \emph{Unseen scenarios} refer to entirely new combinations of context variables that do not appear in the training dataset at all, and hence cannot be directly learned from historical data. Collectively, we refer to the scarcity of data for these infrequent or novel combinations as \emph{context sparsity}. Our proposed method, \textbf{C}ontext \textbf{E}ncoding and \textbf{N}ormalizing \textbf{T}ime \textbf{S}eries Generation, or \textbf{CENTS}, is designed explicitly to tackle context sparsity, enabling high-fidelity generation of synthetic electricity consumption time series even for rare and unseen scenarios.
\par
Using generative models to address dataset shortages has one key downside: The quality of the generative model is determined by the quality of the training data, which means that the quality of the generative model may be held back by the very issue it is trying to address. For that reason, our goal is to design a generative modeling framework that is also capable of extrapolating to entirely new combinations of context variables not seen in the training data. This ensures that our generative model is capable of generating truly \emph{new} synthetic data that enriches the original training dataset. A model only capable of producing the conditions of the training dataset is akin to creating a parametrized sampler of the training data, and is thus a limited way of addressing data shortage. Although CENTS can be used to generate synthetic time series data in any domain, we make design choices that are tailored towards the aforementioned domain-specific issues: It is adaptable to any number of context variables, and can be applied to time-series of arbitrary lengths and dimensionality, This means that it can be used to model electricity consumption and generation jointly, and do so at different granularities (and can thus be extended to longer electricity consumption windows). This makes our method highly adaptable to any dataset. The main capabilities of CENTS are enabled by a context encoder, a flexible embedding module that generates a compressed latent representation of household context variables used to condition generative time series models. This context encoder generates expressive embeddings due to an auxiliary context reconstruction task that is trained jointly with the generative model. This helps the model in reliably identifying infrequently appearing context variables during training, which improves the quality of the generated time series across the dataset. It also helps the model produce highly plausible time series for completely novel contexts.
\par
\textbf{In summary, our contributions are}:
\begin{itemize}
    \item We introduce CENTS, which consists of a context encoder and context normalizer that can be combined with any generative time-series model architecture capable of producing high-quality synthetic time series data for any number and combination of context variables.
    \item To address the task of generating data for very infrequent and completely novel combinations of context variables, we propose a context reconstruction task that can be trained jointly with any generative model, making context embeddings more expressive and improving overall generative model performance.
    \item We provide a comprehensive overview of different evaluation methods for generative time series models.
\end{itemize}

\label{sec:syn_ts_gen}
\section{Problem Statement}

There are two commonly encountered issues when working with time series data in the energy space. We define the first one as \textit{temporal} sparsity, which describes settings with limited data availability along the time axis, i.e. the data availability of household electricity consumption in December. Incorporating temporal information like the month or weekday of the time series into the set of context variables allows us to generate data for a certain time period, alleviating this issue.
\par
We term the second issue as \textit{context} sparsity, which refers to datasets where combinations of context variables appear very infrequently and generate electricity consumption patterns distinct from other contexts. This problem is exacerbated as the number context variables and the number of categories per context variable increases. In a context sparse scenario, it is very difficult for a generative model to learn disentangled representations of individual context variables, because effects of individual context changes cannot be isolated. This makes it very difficult for a generative model to extrapolate to completely new contexts, which is a requirement for being able to synthesize new data. Tackling this issue requires a model design that encourages disentanglement of individual context variable representations that enable the model to generate plausible data for new contexts without hurting the overall generative performance of the model. To avoid any confusion, we want to emphasize the distinction between the problem statement of \textit{generative time series modeling}, which is what we address in this work, and \textit{time series forecasting}. Time series forecasting models are usually autoregressive models, where the forecast is conditioned on time series values of previous time steps. This contrasts with generative time series modeling, where new time series can be generated without requiring past data. We encapsulate this distinction in three rules for valid generative time series modeling:

\begin{itemize}
    \item A generative time series model should generate synthetic time series without explicitly requiring an input of real time series data. 
    \item It should allow for conditional generation.
    \item It should support random sampling, allowing users to generate time series data that reflects the variability inherent in the original data distribution.
\end{itemize}

These rules specify only what we would consider \textit{valid} generative time series modeling, determining what constitutes a \textit{good} generative time series model is another question, discussed further in Section~\ref{sec:eval}.

\section{Related Work}
\label{sec:related_work}
\subsection{Time Series Generation}

Generative modeling of time series data has evolved rapidly in the last decade; initially driven by GANs as in \cite{yoon2019time, Lin2020doppelganger}, and improved with energy models \cite{Jarrett2021}.
generative modeling of time series data has also exploited recent advances in neural network architectures; e.g. Ordinary differential equation networks, which naturally model the temporal dynamics in \cite{Zhou2023}; and Transformers in \cite{Feng_Miao_Zhang_Zhao_2024}.  
More recently, novel diffusion-based approaches such as \cite{Narasimhan2024} and \cite{yuan2024diffusionts} have outperformed GANs in generative modeling tasks. Finally, the arrival of large language models has also had an impact on the time series domain; a survey on the subject was introduced by \cite{Zhang2024llmts}.

\subsection{Synthetic Data for Energy Applications}
Several works highlight the difficulty in obtaining publicly accessible datasets for analyzing electrical consumption in residential buildings. 
Challenges in synthetic energy generation or consumption data are typically: 
1) capturing realistic spatial-temporal correlations among a group of loads served by the same distribution,
2) limited training samples for generating synthetic load profiles, which turns into lack of usable data,
3) lack of metrics that measure realism of synthetic data, 
4) synthetic data lacking high-frequency variations seen in real data, making it less useful for certain applications, 
5) privacy considerations, and
6) biases in Model-Based Assumptions.
%
%
\cite{Razghandi2024} introduce a method combining Variational Auto-Encoder and Generative Adversarial Network techniques to generate time-series data of energy consumption in smart homes. 
\cite{Li2024} use multi-head self-attention mechanisms and prior conditions to capture unique temporal correlations in charging dynamics, and denoising diffusion models to produce realistic charging demand profiles.
\cite{Hu2024} introduce the Multi-Load Generative Adversarial Network (MultiLoad-GAN) for generating synthetic load profiles with realistic spatial-temporal correlations; and two complementary metrics for evaluating the generated load profiles.
\cite{Charbonnier2024} present the Home Electricity Data Generator, an open-access tool for generating realistic residential energy data that leverages GANs and a Markov chain mechanism for successive days, based on the characterization of behavior cluster and profile magnitude transitions.
\cite{Liang2024} develop an Ensemble of Recurrent Generative Adversarial Network to generate high-fidelity synthetic residential load data. They also introduce a loss function that takes into consideration both adversarial loss and differences between statistical properties. 
\cite{Claeys2024} improve realism with a 2 phase proposal: 1) a Dynamically Adaptive GAN to construct annual time series of daily consumption, effectively capturing intraday and seasonal dynamics, and 2) a wavelet-based decomposition-recombination technique to create stochastic daily profiles with realistic intraday variations and peak demand behavior.
\citep{Thorve2023} release a unique, large-scale, digital-twin of residential energy-use dataset for the residential sector 
covering millions of households. The data is comprised of hourly energy use profiles for synthetic households, disaggregated into Thermostatically Controlled Loads and appliance use. 
%
To the best of our knowledge \cite{Lin2020doppelganger} and \cite{Narasimhan2024} are among the few works generating synthetic data for a large number of context variables. However, there are important differences between the mentioned works and ours:
1) although their approach is also end-to-end, our context encoder can be combined with any generative model and is modular and highly flexible, 
2) they learn a joint embedding space of context and time series, whereas we have separate embedding spaces,
3) they do not address context sparsity nor generating data for unseen contexts.

\section{Method}
\label{sec:method}
 \begin{figure*}[!thbp]
    \centering
    \includegraphics[width=0.80\textwidth]{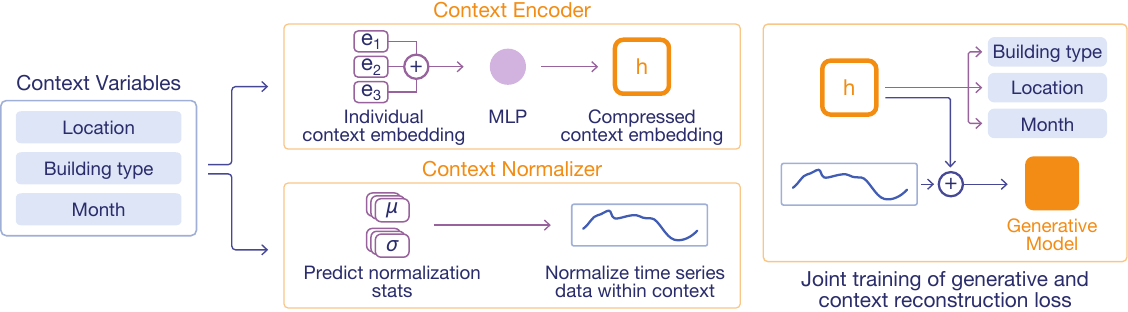}
    \caption{\textbf{CENTS Architecture.} Time series context variables are passed to the context encoder, which generates a compressed context embedding. The context encoder is jointly trained with the generator loss and an auxiliary context reconstruction task that reconstructs the household context from the compressed context embedding. Household context is used to perform within group normalization using a parametric normalizer, and the final household context embedding and the normalized time series are passed to the generative model.}
    \label{fig:method}
\end{figure*}

In this section, we outline the structure and training process of our CENTS, which comprises a context encoder, an auxiliary context reconstruction task and context normalization approach enabling synthesis for arbitrary context combinations. The CENTS architecture is visualized in Figure~\ref{fig:method}.

\subsection{Context Encoder}
\label{subsec:context}

Conditioning the model generation process on a set of context variables $\boldsymbol{C}$ provides the user with increased control over the generated time series data. The goal is to sample generated data $\mathbf{X}_g \sim P(\boldsymbol{X}|\boldsymbol{C})$ that reflects a given context. Our models also need to be able to generate time series for any combination and any number of context variable categories, even those not seen during training. We also want the context encoder to be able to learn disentangled representations of context variables. This ensures that if a previously unseen combination of context variables is passed to the context encoder, the generated embedding is still meaningful and allows the model to extrapolate to new combinations of context variables.

\paragraph{Embedding each Context Variable Individually.}
Let each context variable $c_i$ take values in a finite set of categories. We first map $c_i$ to a continuous \emph{embedding} $\mathbf{e}_i \in \mathbb{R}^{D}$. Concatenating all embeddings yields

\begin{equation}
\mathbf{E} \;=\; [\mathbf{e}_1,\, \mathbf{e}_2,\, \ldots,\, \mathbf{e}_N]
\;\;\in\;\; \mathbb{R}^{N \times D},
\end{equation}
which we  flatten to form a single vector
$\boldsymbol{\varepsilon} \in \mathbb{R}^{N \cdot D}$. Conceptually, $\mathbf{e}_i$ captures information about the variable $c_i$ alone, while $\boldsymbol{\varepsilon}$ unifies these embeddings.

\paragraph{Compressing into a Single Context Embedding.}
To produce a compact representation of a household context, we feed the flattened embeddings $\boldsymbol{\varepsilon}$ into a multi-layer perceptron (MLP) that outputs a final context embedding

\begin{equation}
\mathbf{h} \;=\; \text{MLP}(\boldsymbol{\varepsilon})
\;\in\; \mathbb{R}^{d},
\end{equation}
where $d \ll N \cdot D$ is chosen so that $\mathbf{h}$ is significantly lower dimensional. This forces the embedding to capture only the most meaningful interactions between context variables. 

\paragraph{Auxiliary Reconstruction (Classification) Heads.}
To ensure that $\mathbf{h}$ fully captures $\boldsymbol{C}$, we introduce an auxiliary context reconstruction task - one per context variable - that maps from $\mathbf{h}$ back to a distribution over the context variable categories using another set of MLPs:

\begin{equation}
\widehat{\mathbf{y}}_{i} \;=\; \text{MLP}(\mathbf{h}),
\quad i=1,\ldots,N,
\end{equation}
where $\widehat{\mathbf{y}}_{i} \in \mathbb{R}^{|\mathcal{C}_i|}$ is a logit vector for the $i$-th variable (which has $|\mathcal{C}_i|$ categories). We compute a standard cross-entropy loss comparing $\widehat{\mathbf{y}}_{i}$ to the true category $c_i$. Summing over all context variables yields the \emph{context reconstruction loss}:

\begin{equation}
\mathcal{L}_{\text{aux}} \;=\; \sum_{i=1}^{N}\; \text{CE}\!\bigl(\widehat{\mathbf{y}}_{i},\, c_i\bigr).
\end{equation}

This idea is reminiscent of an autoencoder bottleneck in VAEs \citep{kingma2013auto}, in which high-dimensional inputs (the concatenated embeddings $\boldsymbol{\varepsilon}$) are forced through a smaller hidden embedding space that is trained by reconstructing the original input. This design encourages the model to learn a more disentangled representation of the global context, because individual context variables need to be inferrable from the 
embedding.

\paragraph{Joint Training Objective.}
During training, we have a generative loss $\mathcal{L}_{\text{gen}}$ (e.g., from a GAN or a diffusion model) and the auxiliary context reconstruction loss $\mathcal{L}_{\text{aux}}$. We combine them into a final objective via a scaling factor $\lambda$:

\begin{equation}
\mathcal{L}_{\text{total}} 
\;=\; \mathcal{L}_{\text{gen}} 
\;+\; \lambda\ \cdot \mathcal{L}_{\text{aux}}.
\end{equation}

By adding $\lambda\ \cdot \mathcal{L}_{\text{aux}}$ to the overall training loss, we encourage expressivity and disentanglement of the context embeddings: the model is rewarded for correctly classifying each context variable from $\mathbf{h}$, forcing $\mathbf{h}$ to retain enough information about each variable individually. The learning task is thus to find an expressive, low-dimensional representation of the global context that allows reconstruction to all context variable categories, while simultaneously being a useful conditioning signal for the generative model. The idea is to avoid a collapse of the context representation where differences in representation are solely driven by slight differences in time series. Empirically, this promotes robustness to rarely seen context combinations and improved extrapolation, increasing overall model performance.

\subsection{Context Normalization}
\label{subsec:normalization}

Many generative architectures benefit from mapping time series measurements $\{ x_t \}$ from their original scale (e.g. kWh) to a normalized space (e.g.\ $x_t \in [0,1]$). In our setting, however, a \emph{global} normalization 
(i.e.\ computing a single mean and variance across all households) is suboptimal because households in the dataset have drastically different usage patterns. \emph{Context normalization} --- normalizing each subset of households with the same context variables $\mathbf{c}$ --- yields better generative performance in our case, because the scale and variation of $\{ x_t \}$ are more homogeneous within each group. Although context-grouped normalization can achieve strong results, it faces a key limitation when the model is asked to generate time series for a new combination of context variables $\mathbf{c}$ that was either too rare or never observed in the training set. In both cases, no robust pre-computed statistics exist, making it unclear how to invert from the model's normalized outputs back to the original kWh scale.

\paragraph{Context Normalization.}
We introduce a parametric normalizer $\mathcal{N}_{\theta}$ that \emph{predicts} the normalization statistics based on input context variables $\boldsymbol{C}$. Concretely, let each time series have $d$ dimensions, and let $\boldsymbol{C}$ be a set of integer-coded context variables. We train a small neural network 
\begin{equation}
\bigl(\hat{\boldsymbol{\mu}},\hat{\boldsymbol{\sigma}}, \hat{\boldsymbol{z}}_{\min}, \hat{\boldsymbol{z}}_{\max}\bigr)
= \mathcal{N}_{\theta}(\boldsymbol{C})
\;\;\in\;\;\mathbb{R}^{d}\times \mathbb{R}^{d}\times \mathbb{R}^{d}\times\mathbb{R}^{d},
\end{equation}
that outputs the predicted mean, standard deviation, and $(z_{\min}, z_{\max})$ bounds. Then, given any time series $\mathbf{x} \in \mathbb{R}^{T \times d}$ and context variables $\boldsymbol{C}$ we do forward and inverse normalization as follows:

\begin{enumerate}
    \item \textbf{Forward (Normalization):} 
    \begin{equation}
       \mathbf{\bar z} = \frac{\mathbf{x} - \hat{\boldsymbol{\mu}}}{\hat{\boldsymbol{\sigma}} + \delta},
       \text{\quad then \quad}
       \mathbf{z}= \frac{\mathbf{\bar z} - \hat{\boldsymbol{z}}_{\min}}{\hat{\boldsymbol{z}}_{\max} - \hat{\boldsymbol{z}}_{\min} + \delta}
       \;\
    \end{equation}
    \item \textbf{Inverse (Denormalization):}
    \begin{equation}
       \mathbf{x} = 
         \Bigl(\mathbf{z}\,\bigl(\hat{\boldsymbol{z}}_{\max} - \hat{\boldsymbol{z}}_{\min}\bigr) + \hat{\boldsymbol{z}}_{\min}\Bigr)
         \;(\hat{\boldsymbol{\sigma}} + \delta)\;+\;\hat{\boldsymbol{\mu}}
       \;\
    \end{equation}
\end{enumerate}

Here, $\delta$ is a small constant (e.g.\ $10^{-5}$) to avoid division by zero. By \emph{learning} normalization statistics, the model can handle unseen combinations at inference, providing valid and plausible normalization parameters for new contexts.

\paragraph{Training the Normalizer.}
During training, we collect true statistics $(\mu,\sigma,z_{\min},z_{\max})$ for each observed context group. Then, we build a supervised dataset of $(\mathbf{c},\, \mu,\, \sigma,\, z_{\min},\, z_{\max})$ pairs. We optimize the MSE objective

\begin{align}
\min_{\theta}\;\; \mathbb{E}\Bigl[
   &\bigl\|\hat{\boldsymbol{\mu}} - \mu\bigr\|^2 
   + \bigl\|\hat{\boldsymbol{\sigma}} - \sigma\bigr\|^2 \notag \\
   &+ \bigl\|\hat{\boldsymbol{z}}_{\min} - z_{\min}\bigr\|^2 
   + \bigl\|\hat{\boldsymbol{z}}_{\max} - z_{\max}\bigr\|^2
\Bigr],
\end{align}
which encourages the network to accurately predict the per-dimension statistics from each set of context variables $\boldsymbol{C}$. After convergence, the context normalizer can transform or invert any sample given arbitrary context variable combinations. Intuitively, it does so by approximating the required normalization statistics based on the statistics of known context variable combinations that are very similar, e.g. where only one or two context variables have changed. In practice, we found that training a relatively shallow network is sufficient for generating reliable predictions.

\subsection{Generative Models}
\label{subsec:models}

In the following section, we briefly outline the different generative model architectures we combine with CENTS. We want to emphasize that CENTS can be used in conjunction with any generative model architecture. We thus start with a generalized description of how the context embedding $\mathbf{h}$ is used to condition the training process of a generative model.
\par
Let $\mathbf{z} \in \mathbb{R}^{B \times T \times D}$ be a random noise tensor where $B$ is the batch size, $T$ is the time series length and $D$ is the dimensionality of the time series, and let $\mathbf{h} \in \mathbb{R}^d$ be the context embedding from our context encoder. To integrate $\mathbf{h}$ with $\mathbf{z}$, we repeat $\mathbf{h}$ across dimension $T$ to form $\mathbf{h}^\prime \in \mathbb{R}^{B \times T \times d}$. We concatenate $\mathbf{z}$ and $\mathbf{h}^\prime$ along the feature dimension, yielding:

\begin{equation}
\mathbf{z}^\star \;=\; 
\bigl[\mathbf{z},\;\mathbf{h}^\prime\bigr]
\;\in\;\mathbb{R}^{B \times T \times (D + d)}.
\end{equation}

A generator $\mathcal{G}_{\theta}$ then consumes $\mathbf{z}^\star$ to produce a synthetic time series ${\mathbf{X}_g} \in \mathbb{R}^{B \times T \times D}$; such that 
${\mathbf{X}_g} = \mathcal{G}_{\theta}\!\bigl(\,\mathbf{z}^\star\bigr). $
%
During \emph{training}, real data is paired with random $\mathbf{z}$ and the corresponding context vectors to optimize the generative objective. At \emph{inference}, we again sample noise $\mathbf{z}$, replicate the context embedding $\mathbf{h}$, concatenate, and feed the result through $\mathcal{G}_{\theta}$ to synthesize the new time series. The models chosen here are GANs \citep{goodfellow2014generative} and diffusion models \citep{ho2020denoising}, being the most popular generative modeling paradigms today. Thus, we select a simple baseline model and two specialized time series architectures to evaluate CENTS. 
\paragraph{ACGAN.} 
In this GAN architecture \citep{odena2017acgan}, the generator is conditioned on a set of class labels, and the discriminator is not only trained to distinguish between the real and generated time series, but also classifies the data into correct class labels. This dual objective helps improve the training of both the generator and the discriminator, leading to higher-quality generated samples. Although GANs have been widely adopted for generative modeling, they frequently suffer from issues around mode collapse and unstable training, due to the adversarial training dynamics of the generator and the discriminator. We use CENTS with an ACGAN \citep{odena2017acgan} architecture that supports multi-dimensional time series generation and embedding-based conditioning.

\paragraph{\textbf{Diffusion-TS.}}
Diffusion-TS \citep{yuan2024diffusionts} is a recent diffusion modeling framework capable of generating novel multivariate time series samples using a transformer-based encoder-decoder architecture that captures disentangled temporal representations of the input time series; therefore highly suitable to our task. Each decoder block consists of a transformer layer, and a series of ``interpretable'' layers that introduce inductive biases to encode specific semantic information. These are: a trend layer designed to capture the low-frequency, smooth movements of the time series, and a  Fourier synthetic layer for high-frequency periodic patterns. 

\section{Evaluation Metrics}
\label{sec:eval}
We distinguish between metrics that capture the \textit{fidelity} of the generated time series, and metrics that capture the \textit{utility} of the generated time series. The fidelity of a time series describes how closely the generated data resembles the real data, whereas utility describes how useful the generated data is for downstream tasks (like e.g., forecasting and anomaly detection) compared to the real data. Although it is likely that there is a strong correlation between the fidelity and utility of generated time series, we make this distinction to clarify the evaluation methodology in this work and to highlight that there may be cases where those metrics do not correlate perfectly.

\subsection{Fidelity Metrics}

Many commonly employed fidelity metrics are defined at the time series level. Given a synthetically generated time series $\mathbf{X}_g$  and a corresponding real time series $\mathbf{X}_r$, the following metrics measure how closely $\mathbf{X}_g$ resembles $\mathbf{X}_r$.

\paragraph{Multivariate Dynamic Time Warping Distance (MDTWD).}
Extending Dynamic Time Warping (DTW) to multivariate time series \citep{shifaz2023elasticsimilaritydistancemeasures}, MDTWD aligns two sequences $\mathbf{X}_g$ and $\mathbf{X}_r$, with same  length $T$ and dimension $d$, finding a path that minimizes the cumulative per-timestep squared distance, subject to DTW constraints ensuring valid alignments. 
%



%
%
\paragraph{Maximum Mean Discrepancy (MMD)} is a statistical measure that quantifies the similarity between real and synthetic data. 

%
\paragraph{Banded Mean Squared Error (BMSE)}
checks if synthetic values lie outside the per-timestep range of real data. 
%
It penalizes large deviations outside the observed real range; thus, it is  a good proxy for the overall plausibility of the synthetic time series, irrespective of fine-grained shape.

\subsection{Utility Metrics}

Fidelity metrics capture similarity to real data but do not necessarily measure how well synthetic data can replace real data in downstream tasks. Utility assesses how effectively synthetic data can be used for other machine learning tasks.

\paragraph{Discriminative Score}\citep{yoon2019time}.
Instead of training a downstream task on synthetic data and evaluating on real, this metric trains a binary classifier to distinguish real vs. synthetic examples. Its final accuracy is usually offset by $0.5$. A discriminative score near $0$ thus means real and synthetic samples are practically indistinguishable by a classifier.

\paragraph{Predictive Score}\citep{yoon2019time}.
Here, a generic time-series forecasting model is trained on synthetic data and tested on real data. Let $\{\hat{y}_{i,t,c}\}$ be predictions on real time series $\{y_{i,t,c}\}$, then we compute the mean absolute error (MAE) across all $N$ samples, $T$ time steps, and $d$ dimensions:

\begin{equation}
\text{Pred. Score} 
= \frac{1}{N}\,\sum_{i=1}^{N}\,\frac{1}{T}\,\sum_{t=1}^{T}\,\frac{1}{d}\,\sum_{c=1}^d
  \bigl|\hat{y}_{i,t,c} - y_{i,t,c}\bigr|.
\end{equation}

A lower score implies better alignment between model predictions and the real series, indicating higher utility of the synthetic dataset for forecasting tasks.

\textbf{Context-FID} \citep{jeha2022psa} adapts the Fréchet Inception Distance (FID) \citep{heusel2017gans} commonly applied in computer vision to time series. The idea is to compute embeddings of the real time series and the synthetic time series datasets using a time series embedding model trained on the original data, and then to compute the Fréchet distance between the real time series dataset embedding and the synthetic time series dataset embedding. Context-FID is based on TS2Vec \citep{yue2022ts2vec}, a self-supervised framework for learning rich time series representations. We generally find that Context-FID seems to be most in-line with visual evaluations of the quality of the synthetic data.




\section{Experiments}
\label{sec:exp}
\begin{table*}[h!]
\centering
\resizebox{\textwidth}{!}{%
\begin{tabular}{l c c c c c c c c c}
\toprule
\textbf{Model} & \multicolumn{2}{c}{\textbf{Context-FID $\downarrow$}} & \multicolumn{2}{c}{\textbf{BMSE $\downarrow$}} & \multicolumn{2}{c}{\textbf{MMD $\downarrow$}} & \multicolumn{2}{c}{\textbf{MDTWD $\downarrow$}} \\
\cmidrule(lr){2-3} \cmidrule(lr){4-5} \cmidrule(lr){6-7} \cmidrule(lr){8-9}
 & Univariate & Multivariate & Univariate & Multivariate & Univariate & Multivariate & Univariate & Multivariate \\
\midrule
GAN Baseline ($\lambda=0$) & 4.971 & 4.802 & 1.167 & 0.438 & 0.263 & 0.355 & 15.062 & 15.748 \\
GAN Baseline ($\lambda=0.1$) & 2.333 & 4.956 & 0.603 & 0.435 & 0.281 & 0.333 & 10.864 & 14.848 \\
ACGAN ($\lambda=0$) & 0.477 & 3.454 & 0.110 & 0.247 & 0.142 & 0.296 & 8.745 & 16.326 \\
ACGAN ($\lambda=0.1$) & 0.474 & 3.735 & 0.1374 & 0.238 & 0.145 & 0.304 & 8.700 & 15.115 \\
Diffusion-TS ($\lambda=0$) & \underline{0.473} & \underline{0.792} & \underline{0.006} & \underline{0.008} & \textbf{0.086} & \textbf{0.109} & \underline{6.759} & \underline{10.494} \\
Diffusion-TS ($\lambda=0.1$) & \textbf{0.24} & \textbf{0.735} & \textbf{0.005} & \textbf{0.008} & \underline{0.089} & \underline{0.111} & \textbf{6.719} & \textbf{10.465} \\
\bottomrule
\end{tabular}%
}
\caption{Evaluation metrics for GAN Baseline, ACGAN, and Diffusion-TS models across univariate and multivariate synthetic time series. Diffusion-TS with $\lambda=0.1$ consistently performs best.}
\label{tab:metrics_comparison}
\end{table*}

\begin{table}[t!]
\centering
\begin{tabular}{l c c }
\toprule
\textbf{Model} & \textbf{Pred. Score $\downarrow$} & \textbf{Discr. Score $\downarrow$} \\
\midrule
GAN Baseline ($\lambda=0$) & \underline{0.395} & 0.258  \\
GAN Baseline ($\lambda=0.1$) & 0.402 & 0.299  \\
ACGAN ($\lambda=0$) & 0.392 & 0.121 \\
ACGAN ($\lambda=0.1$) & 0.403 & 0.116 \\
Diffusion-TS ($\lambda=0$) & \underline{0.395} & \underline{0.099}\\
Diffusion-TS ($\lambda=0.1$)  & \textbf{0.388} & \textbf{0.037}\\
\bottomrule
\end{tabular}
\caption{Utility metrics correlate with Context-FID. Diffusion-TS is best-performing model.}
\label{tab:utility_metrics}
\end{table}

\begin{table}[h!]
\centering
\begin{tabular}{l c cc}
\toprule
\textbf{Model} & $\boldsymbol{\lambda}$
& \multicolumn{2}{c}{\textbf{Context-FID} $\downarrow$} \\
\cmidrule(lr){3-4}
 & &  Overall & Sparse-only \\
\midrule
GAN Baseline & 0.0 
& 4.971 & 5.765 \\
GAN Baseline & 0.1 
& 2.333 & 5.300 \\
ACGAN & 0.0 
& 0.477 & 2.491 \\ 
ACGAN & 0.1 
& 0.474 & 1.346 \\
Diffusion-TS & 0.0 
& \underline{0.473} & \underline{0.802} \\
Diffusion-TS & 0.1 
& \textbf{0.240} & \textbf{0.467} \\
\bottomrule
\end{tabular}
\caption{Context reconstruction loss weights of $\lambda > 0$ improve generative performance across the board, driven especially by improvements on sparse context samples.}
\label{tab:lambda_ablation}
\end{table}

\textbf{PecanStreet Dataport Dataset.} 
We use the Pecan Street Dataport dataset \citep{2025dataport} for our main experiments. This dataset provides household-level electric load and photovoltaic (pv) generation time series, as well as comprehensive metadata for 75 households spread across California, New York and Texas. The time series data is given in kWh and represents the net electricity imported (or exported if generation exceeds usage) from the public grid. Residential load and pv generation data is available at 15 minute granularity, for time periods of 6-12 months depending on the household. The household metadata contains additional information like building types, building square footage, building construction year and more fine-grained location data. We aggregate the data to daily time series load profiles consisting of 96 values. This amounts to a total of $21578$ daily load profiles. Each load profile is paired with household metadata, and we include both month and weekday of that load profile as additional context, resulting in $9$ context variables. Around half of the households also have photovoltaics (pv), and electricity generation time series data in kWh is available for those households at a similar granularity. We use this subset of pv-owning households for our multi-dimensional time series experiments.

\begin{figure}[h!]
    \centering
    \includegraphics[width=1.0\linewidth]{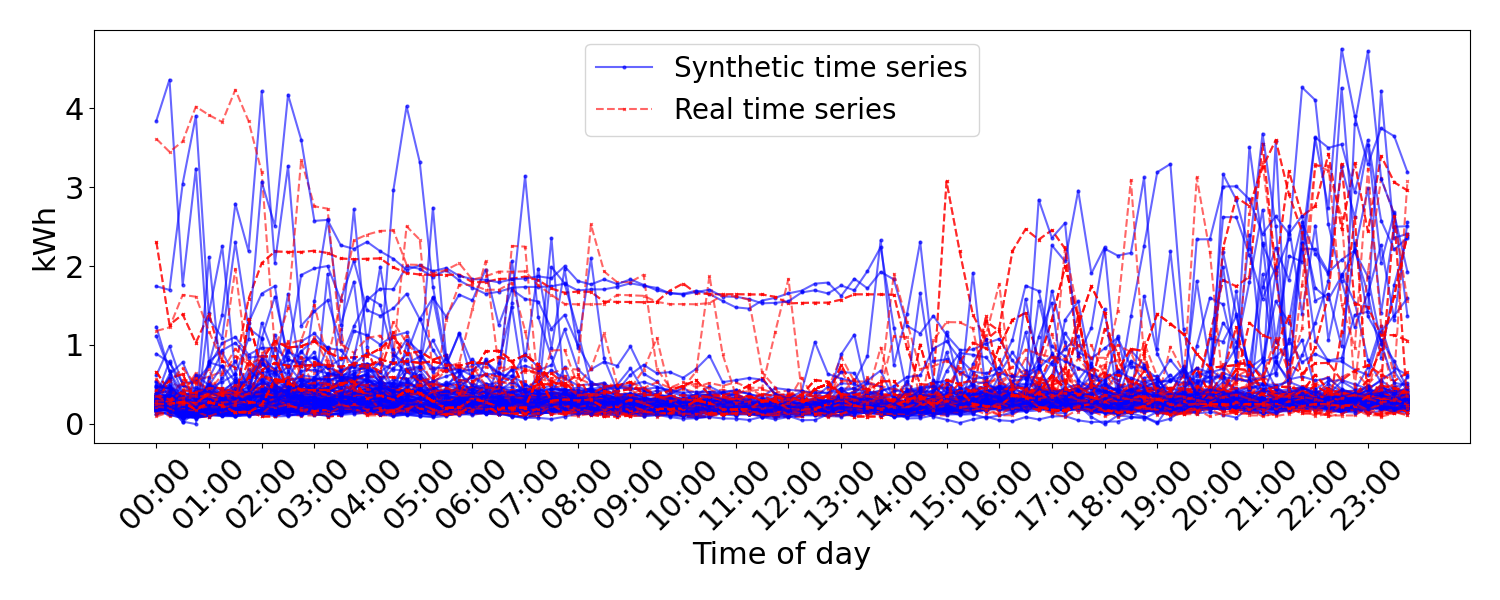}
    \caption{CENTS with Diffusion-TS captures multi-modal distribution of real time series shapes. Shows a comparison between 100 synthetic time series (blue) and the corresponding closest real time series (red) for the same context.}
    \label{fig:diff_syn}
\end{figure}

\begin{figure}[h!]
    \centering
    \includegraphics[width=1.0\linewidth]{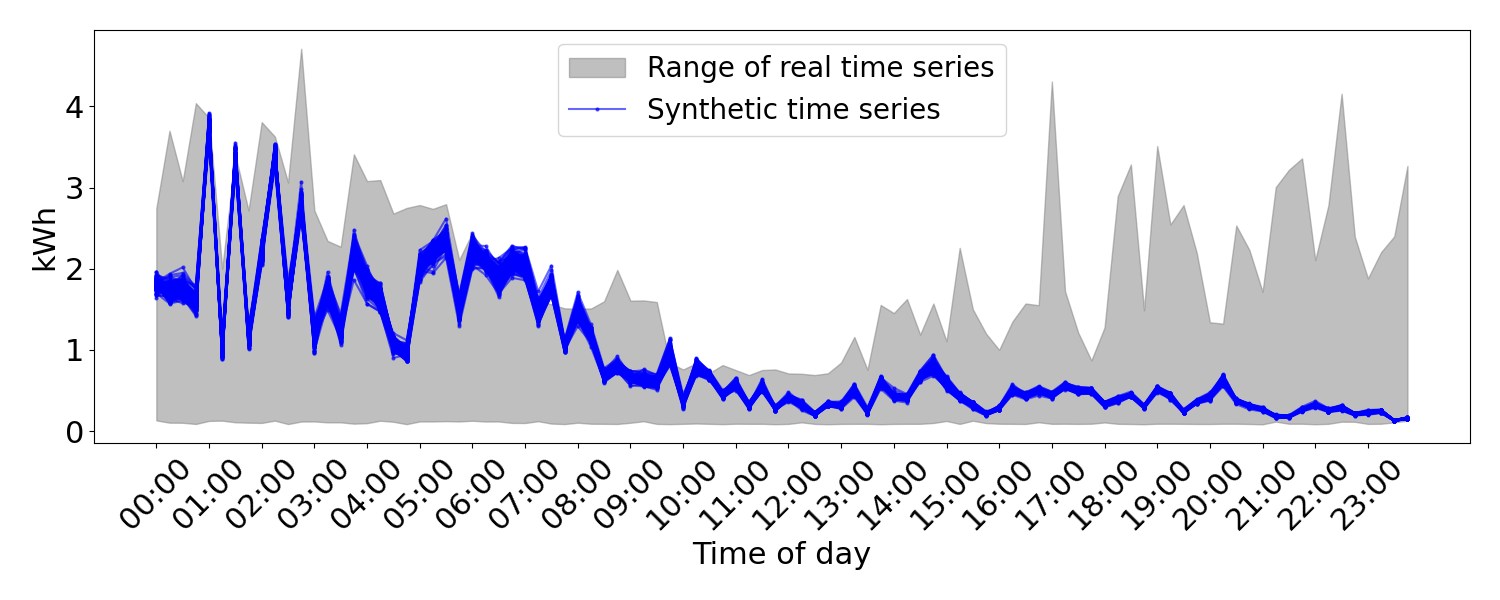}
    \caption{CENTS with ACGAN is able to produce highly plausible synthetic time series within dataset bounds, but shows lower variance of time series shapes.}
    \label{fig:acgan_range}
\end{figure}

\paragraph{\textbf{Evaluating Generative Performance.}} Our experiments demonstrate three key properties of our models: First, all models generate high-fidelity synthetic time series that effectively capture the real data distribution. Second, the context reconstruction loss term in the CENTS framework significantly improves overall model performance, particularly for sparse context time series. Third, CENTS can extrapolate to generate plausible time series for unseen contexts. We compare the architectures outlined in Section~\ref{subsec:models} with a vanilla GAN baseline. Metrics are obtained by generating synthetic equivalents of the training dataset using models trained under similar conditions on identical hardware (details in Appendix). This synthetic dataset is evaluated via the metrics outlined in Section~\ref{sec:eval}. Fidelity metrics are averaged across all time series in the synthetic dataset. Results are presented in Table~\ref{tab:metrics_comparison} and Table~\ref{tab:utility_metrics}. Both ACGAN and Diffusion-TS substantially outperform the GAN baseline across utility and fidelity metrics. We also find that this holds across univariate and multivariate generation, although overall model performance decreases for multivariate generation. The performance improvement of adding the context reconstruction loss is substantial for Diffusion-TS and the baseline, but only marginal for ACGAN. This is likely because the ACGAN discriminator and generator use a binary cross entropy classification loss that may already enforce a level of disentanglement between context variable representations. This is supported by the fact that the auxiliary reconstruction task does substantially improve baseline Context-FID. Figures \ref{fig:diff_syn} and \ref{fig:acgan_range} illustrate the different capabilities of CENTS with Diffusion-TS and CENTS with ACGAN. While ACGAN is able to produce high fidelity time series (i.e. plausible time series within the range of values seen during training), Diffusion-TS excels at mode coverage, exhibiting larger variance of generated time series that are still true to real data.

\paragraph{\textbf{Ablating Context Encoder Context Reconstruction Objective.}} To evaluate the importance of the context reconstruction objective in CENTS, we set the loss weight to $\lambda = 0$ and compare these models to those with $\lambda > 0$. Note that we observe fast convergence of the context reconstruction loss for all $\lambda > 0$, which is why we do not ablate the exact choice of $\lambda$ further. To analyze the objective's impact on context-sparse time series generation specifically, we require a ground-truth label for context sparsity. We define proxy labels for sparsity using a combination of two measures: (1) Frequency-based sparsity labels identify context combinations not within the 90th percentile of most frequent combinations in our dataset. (2) Clustering-based sparsity labels are derived by computing summary statistics (mean, std, min, max) of all time series, clustering with KMeans, and labeling the smallest 10\% of clusters by count as sparse. The final sparsity label is the union of these two sets. Using these proxy labels, we compute the metrics from Section~\ref{sec:eval} separately for sparse and non-sparse contexts. Empirical results indicate a significant Context-FID drop-off for sparse samples across all models, validating this approach as a useful proxy for identifying cases where synthetic data generators struggle. Our results in Table~\ref{tab:lambda_ablation} show that our context reconstruction objective substantially improves Context-FID on context-sparse samples. Due to the low number of context sparse time series, this does not always substantially impact overall Context-FID, but underlines the fact that our context encoder improves performance for very infrequently appearing contexts specifically.

\paragraph{\textbf{Extrapolating to Unseen Contexts.}}

\begin{figure}[!htbp]
    \centering
    \begin{subfigure}[t]{0.80\linewidth}
        \centering
        \includegraphics[width=\linewidth]{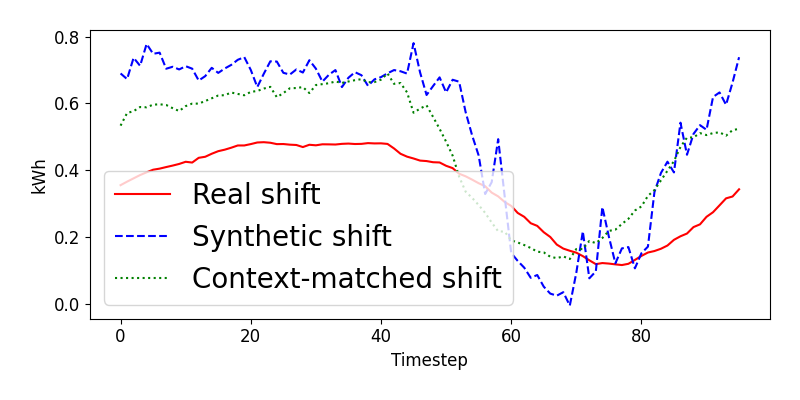}
    \end{subfigure}
    \begin{subfigure}[t]{0.80\linewidth}
        \centering
        \includegraphics[width=\linewidth]{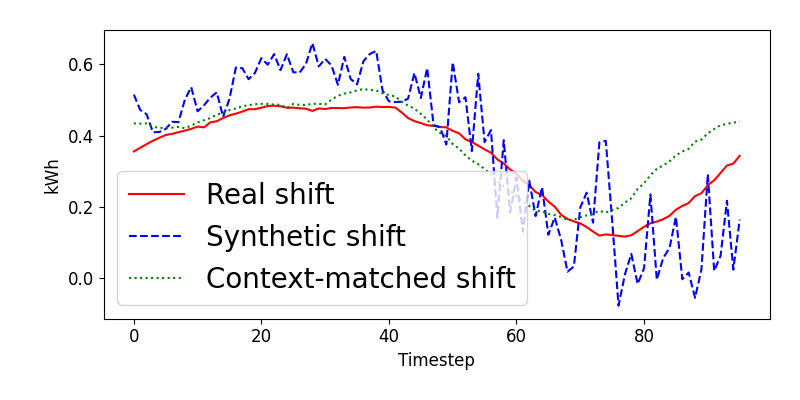}
    \end{subfigure}
    \caption{Shows the mean per-timestep kWh difference over the dataset (red), in-context difference (same location and building type; green), and the model's extrapolated shift in consumption (blue) for Diffusion-TS ($\lambda=0.1$) (top) and ACGAN ($\lambda=0.1$) (bottom). The model's extrapolated shift is less smooth because it is the difference between two synthetic time series, whereas the real and context-matched shift are averages computed over the dataset (or a subset comprising many time series).}
    \label{fig:shift_comparison}
\end{figure}

To assess our model's ability to extrapolate to completely new contexts, we evaluate our model's capabilities to learn disentangled representations of context variables by focusing on the transition from households without photovoltaics (pv) to households with pv. This corresponds to a real-world scenario of interest: understanding how electricity consumption changes when a household installs solar panels. Evaluating the quality of a time series conditioned on an entirely new combination of context variables is difficult, because there is no ground-truth to compare the synthetic data to. We can however, compute an ``average shift'' in consumption patterns from households with pv to those without. Intuitively, if our model is able to disentangle the impact of pv on time series patterns, the difference in generated time series for a household with and without pv should correspond to this average shift across all contexts. To evaluate this, we select all context variable combinations that are either missing households with or without pv. For a randomly sampled combination of context variables, we generate both scenarios and compare the per-timestep shift with both the average shift across the dataset and a context-matched shift (which computes the average per timestep consumption difference for households in the same location and building type). Figure~\ref{fig:shift_comparison} shows that CENTS with ACGAN and Diffusion-TS is able to generate this shift adequately for an individual household, presenting evidence that our models are able to extrapolate proficiently.

\section{Conclusion}
\label{sec:conc}
While other domains, such as language and vision, seem to be running out of data very gradually, in the energy domain, data availability issues are prevalent right from the beginning. In our work, we have shown that CENTS can synthesize high-quality time series data for sparse contexts as well as completely new contexts, and works well with both GAN and diffusion models. We see substantial potential in scaling our ideas up to larger datasets comprised of many different sources with a much broader range of context variables. We are also aware that evaluating the quality of synthetic data for completely new contexts is a difficult task, and that our evaluation approach of this capability is only a first step in this direction.
\newpage

\bibliographystyle{named}
\bibliography{ref}

\clearpage
\section{Appendix}
\subsection{ACGAN Loss Functions and Architecture}

The loss functions used in the ACGAN \citep{odena2017acgan} implementation are designed to optimize the generator $G$ and discriminator $D$ jointly, while incorporating auxiliary classification losses for conditioning variables. These loss terms ensure that $G$ generates realistic and contextually accurate synthetic time series, and that $D$ accurately distinguishes real from fake data while predicting the correct conditioning labels.

\subsubsection{Discriminator Loss}
The discriminator loss $\mathcal{L}_D$ consists of an adversarial loss and auxiliary classification losses:
\begin{equation}
\begin{aligned}
\mathcal{L}_D = & -\mathbb{E}_{x \sim p_{\text{data}}(x)} \left[ \log D(x) \right] \\
& -\mathbb{E}_{z \sim p_z(z), c \sim p(c)} \left[ \log \left( 1 - D(G(z, c)) \right) \right] \\
& + \gamma \sum_{i=1}^{N} \Big( \mathcal{L}_{\text{BCE}}(c_i, \hat{c}_i^{(\text{real})}) + \mathcal{L}_{\text{BCE}}(c_i, \hat{c}_i^{(\text{fake})}) \Big),
\end{aligned}
\end{equation}
where:
\begin{itemize}
    \item $D(x)$ represents the discriminator output for real data.
    \item $G(z, c)$ is the generator output for noise $z$ and context variables $c$.
    \item $\mathcal{L}_{\text{BCE}}$ is the binary cross-entropy loss for auxiliary classification, ensuring $D$ predicts the correct class labels $c_i$ for both real ($\hat{c}_i^{(\text{real})}$) and fake ($\hat{c}_i^{(\text{fake})}$) data. Note that this is a binary cross-entropy loss, and this different from our context reconstruction task in the context encoder.
    \item $\gamma$ is the weighting parameter for the auxiliary classification loss. 
\end{itemize}

\subsubsection{Generator Loss}
The generator loss $\mathcal{L}_G$ combines adversarial and auxiliary classification objectives. It is given by:
\begin{equation}
\begin{aligned}
\mathcal{L}_G = & -\mathbb{E}_{z \sim p_z(z), c \sim p(c)} \left[ \log D(G(z, c)) \right] \\
& + \lambda \sum_{i=1}^{N} \mathcal{L}_{\text{BCE}}(c_i, \hat{c}_i^{(\text{fake})}),
\end{aligned}
\end{equation}
where $\lambda$ controls the contribution of the auxiliary classification loss.

\subsubsection{Model Architecture}
The ACGAN consists of a generator $G$ and a discriminator $D$, both implemented with 1D transposed convolutional layers and auxiliary classification heads.

\subsubsection{Generator}
The generator takes as input a noise vector $z \in \mathbb{R}^{256}$ and (optionally) context embeddings as input. It upsamples the input through transposed convolutional layers, producing time series data of shape $(\text{batch size}, \text{sequence length}, \text{time series dimensions})$. The key architectural components are:
\begin{itemize}
    \item Fully connected layer projecting $z$ and conditioning embeddings to a flattened representation.
    \item Three transposed convolutional layers with feature map dimensions halving at each step: $256 \to 128 \to 64 \to 1$.
    \item LeakyReLU activations and batch normalization at intermediate layers.
    \item Sigmoid activation at the final layer to produce normalized outputs.
\end{itemize}

\subsubsection{Discriminator}
The discriminator uses convolutional layers to downsample input time series data, with an auxiliary classification head for each conditioning variable. The architecture includes:
\begin{itemize}
    \item Three 1D convolutional layers with increasing feature map dimensions: $1 \to 64 \to 128 \to 256$.
    \item Fully connected layer for binary real/fake classification.
    \item Auxiliary classifiers implemented as fully connected layers for each conditioning variable.
    \item LeakyReLU activations and batch normalization in intermediate layers.
\end{itemize}

The total number of trainable parameters in the ACGAN is approximately \textbf{1.33M}. Our vanilla GAN baseline implementation and architecture is equivalent to ACGAN, with the omission of $\mathcal{L}_{\text{BCE}}$.

\subsubsection{ACGAN Hyperparameters}
The hyperparameters used in training the ACGAN are summarized in Table~\ref{tab:hyperparams}.

\begin{table}[h]
\centering
\begin{tabular}{lc}
\toprule
\textbf{Hyperparameter}          & \textbf{Value} \\
\midrule
Noise dimension                  & 256 \\
Conditioning embedding dimension & 16 \\
Batch size                       & 1024 \\
Sampling batch size              & 4096 \\
Number of epochs                 & 5000 \\
Generator learning rate          & $3 \times 10^{-4}$ \\
Discriminator learning rate      & $1 \times 10^{-4}$ \\
\bottomrule
\end{tabular}
\caption{Hyperparameters used for training the ACGAN.}
\label{tab:hyperparams}
\end{table}

\subsection{Diffusion-TS: Loss Functions and Architecture}

Diffusion-TS \citep{yuan2024diffusionts} introduces an interpretable decomposition-based architecture inspired by seasonal-trend decomposition. The model leverages a transformer-based encoder-decoder architecture, where the encoder extracts global patterns from noisy sequences, and the decoder performs decomposition into trend, seasonality, and error components. 

\subsubsection{Diffusion Framework}
 In the forward diffusion process, data $x_0 \sim q(x)$ is gradually noised to a Gaussian distribution $x_T \sim \mathcal{N}(0, \mathbf{I})$ using transitions $q(x_t | x_{t-1}) = \mathcal{N}(x_t; \sqrt{1 - \beta_t}x_{t-1}, \beta_t\mathbf{I})$, where $\beta_t$ is the noise schedule. In the reverse process, the neural network learns to denoise the sample via transitions $p_\theta(x_{t-1} | x_t) = \mathcal{N}(x_{t-1}; \mu_\theta(x_t, t), \Sigma_\theta(x_t, t))$. 

The learning objective is defined as:
\begin{equation}
    \mathcal{L}(x_0) = \sum_{t=1}^T \mathbb{E}_{q(x_t|x_0)} \left[ \| \mu(x_t, x_0) - \mu_\theta(x_t, t) \|^2 \right],
\end{equation}
where $\mu(x_t, x_0)$ is the posterior mean $q(x_{t-1} | x_0, x_t)$. This objective optimizes a weighted variational lower bound on the log-likelihood of the data.

\subsubsection{Model Architecture}
The architecture incorporates a transformer-based encoder-decoder design with specific layers for disentanglement into trend, seasonality, and error components. The decoder is equipped with interpretable layers, including a trend synthesis layer and a Fourier-based seasonality synthesis layer. Our Diffusion-TS implementation has a total of \textbf{2.9M} trainable parameters.

\subsubsection{Trend Synthesis}
Trend representation captures the smooth, gradually varying components of the time series. It is modeled using a polynomial regressor:
\begin{equation}
    V_{\text{tr}}^t = \sum_{i=1}^D \left( \mathbf{C} \cdot \text{Linear}(w_{\text{tr}}^{i,t}) + \mathcal{X}_{\text{tr}}^{i,t} \right),
\end{equation}
where $\mathbf{C} = [1, c, \dots, c^p]$ is the matrix of polynomial terms, $\mathcal{X}_{\text{tr}}^{i,t}$ is the mean output of the $i$th decoder block, and $p$ is a small degree (e.g., $p=3$).

\subsubsection{Seasonality and Error Synthesis}
Seasonal components are extracted using Fourier synthesis. Seasonal representation is modeled as:
\begin{align}
    A_{i,t}^{(k)} & = \left| \mathcal{F}(w_{\text{seas}}^{i,t})_k \right|, \quad \Phi_{i,t}^{(k)} = \phi(\mathcal{F}(w_{\text{seas}}^{i,t})_k), \\
    S_{i,t} & = \sum_{k=1}^K A_{i,t}^{\kappa_{i,t}^{(k)}} \cos(2\pi f_{\kappa_{i,t}^{(k)}} \tau c + \Phi_{i,t}^{\kappa_{i,t}^{(k)}}),
\end{align}
where $\mathcal{F}$ is the Fourier transform, $A_{i,t}^{(k)}$ and $\Phi_{i,t}^{(k)}$ represent amplitude and phase, and $f_k$ denotes frequency. Top-$K$ frequencies are selected to model significant periodic patterns. The error component captures residuals after removing trend and seasonality.

The reconstructed signal is:
\begin{equation}
    \hat{x}_0(x_t, t, \theta) = V_{\text{tr}}^t + \sum_{i=1}^D S_{i,t} + R,
\end{equation}
where $R$ represents residual noise.

\subsubsection{Training Objectives}
The training objective combines time and frequency domain losses to enhance reconstruction accuracy:

\begin{equation}
\begin{split}
    \mathcal{L}_\theta = \mathbb{E}_{t, x_0} \bigg[ w_t \big( & \lambda_1 \| x_0 - \hat{x}_0(x_t, t, \theta) \|^2 \\
    & + \lambda_2 \| \mathcal{F}(x_0) - \mathcal{F}(\hat{x}_0(x_t, t, \theta)) \|^2 \big) \bigg]
\end{split}
\end{equation}

where $\mathcal{F}$ is the Fourier transform, and $\lambda_1, \lambda_2$ are weights for time and frequency domain losses, respectively.

\begin{table}[h]
\centering
\begin{tabular}{lc}
\toprule
\textbf{Hyperparameter}            & \textbf{Value} \\
\midrule
Noise dimension                    & 256 \\
Conditioning embedding dimension   & 16 \\
Batch size                         & 1024 \\
Sampling batch size                & 4096 \\
Number of epochs                   & 5000 \\
Number of diffusion steps          & 1000 \\
Base learning rate                 & $1 \times 10^{-4}$ \\
Encoder layers                     & 4 \\
Decoder layers                     & 5 \\
Model dimension ($d_\text{model}$) & 128 \\
Attention heads                    & 4 \\
Loss type                          & L1 \\
Beta schedule                      & Cosine \\
EMA decay                          & 0.99 \\
Gradient accumulation steps        & 2 \\
\bottomrule
\end{tabular}
\caption{Hyperparameters for training the Diffusion-TS model.}
\label{tab:diffusion_ts_hyperparams}
\end{table}

\subsection{Training Details}

Since none of our models exceed parameter counts of $3$ million, we are able to train our models on a single Nvidia Titan RTX GPU. We run training for all models for a total of $5000$ epochs at a batch size of $1024$. For our GAN-based generative models, we use a constant learning rate of $3e-4$ for the generator and $1e-4$ for the discriminator. For Diffusion-TS, we use a minimum learning rate of $1e-5$, as well as a cosine beta schedule and $1000$ sampling steps. Note that we do not spend significant amounts of time tuning these particular hyperparameters and that improvements in model performance are certainly possible when fine-tuning these further.

\subsection{Evaluation Metrics}

For the sake of clarity and completeness, we include here the mathematical expression of the fidelity metrics. 

\paragraph{Multivariate Dynamic Time Warping Distance.}

Let $x_t^{g} \in \mathbb{R}^d$ and $x_t^{r} \in \mathbb{R}^d$ be the respective feature vectors at time $t$. The per-timestep distance is
\begin{equation}
    d\bigl(x_t^{g}, x_t^{r}\bigr) \;=\; \sum_{i=1}^{d}\,\bigl(x_{t}^{g,i} - x_{t}^{r,i}\bigr)^2.
\end{equation}
MDTWD finds a path $\pi$ over time steps minimizing the cumulative distance:

\begin{equation}
\text{MDTWD}(X_g, X_r) \;=\; \min_{\pi}\,\sum_{(i,j)\in \pi}\,d(x_i^g,\;x_j^r),
\end{equation}
subject to DTW constraints ensuring valid alignments. 
For a dataset of $n$ real-synthetic pairs, the average MDTWD is
\begin{equation}
\text{MDTWD}_{\text{avg}} \;=\; \frac{1}{n}\,\sum_{i=1}^{n}\,\text{MDTWD}\!\bigl(\hat{X}_i,\;X_i\bigr),
\end{equation}
where each synthetic $\hat{X}_i$ pairs with a real $X_i$ sharing the same context variables.

\paragraph{Maximum Mean Discrepancy.}

Given real time series samples $\mathbf{X}_r = \{y_{t}\}^{T}_{t=1}$ and synthetic samples $\mathbf{X}_g = \{x_{t}\}^{T}_{t=1}$, MMD is defined as:
\[
\text{MMD} = \Bigg\{ 
\frac{1}{N^2} \sum_{i=1}^{N} \sum_{j=1}^{N} K(x_{i}, x_{j}) 
- \frac{2}{MN} \sum_{i=1}^{N} \sum_{j=1}^{M} K(x_{i}, y_{j})
\]

\begin{align}
+ \frac{1}{M^2} \sum_{i=1}^{M} \sum_{j=1}^{M} K(y_{i}, y_{j}) 
\Bigg\}^{1/2},
\end{align}
where $K$ is a Gaussian radial basis function kernel, given by $K(x, y) = \exp\left( -\frac{\|x - y\|^2}{2\sigma^2} \right)$. Lower values indicate greater similarity. 

\paragraph{Banded Mean Squared Error.}

For each timestamp $t$, let $B= (b^l_t,b^u_t)$ be the interval between  min and max real values with matching conditioning. 
The BMSE for synthetic $\mathbf{X}_g$ is:
\begin{equation}
\mathrm{BMSE}(X_g,B) \;=\; \frac{1}{T}\,\sum_{t=1}^T\,\min\!\Bigl(\bigl(x_t - b^l_{t}\bigr)^2,\,\bigl(x_t - b^u_{t}\bigr)^2\Bigr).
\end{equation}

\paragraph{Context-FID}

is given by:
\begin{equation}
\text{Context-FID} = \|\mu_r - \mu_g\|^2 + \text{Tr}\left( \Sigma_r + \Sigma_g - 2 \sqrt{\Sigma_r \Sigma_g} \right)
\end{equation}
where:
\(\mu_r\) and \(\Sigma_r\) are the mean and covariance of the real data's embeddings, \(\mu_g\) and \(\Sigma_g\) are the mean and covariance of the generated data's embeddings and \(\text{Tr}(\cdot)\) denotes the trace operator, which is equivalent to computing the sum of the diagonal elements of a matrix. 

\subsection{Further Qualitative Results}

\begin{figure}[h!]
    \centering
    \includegraphics[width=0.9\linewidth]{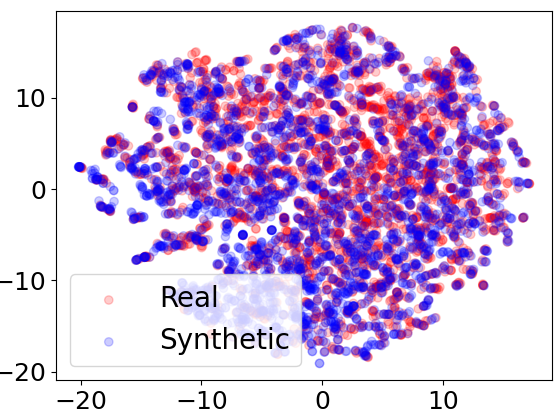}
    \caption{TSNE \citep{van2008visualizing} visualization of synthetic and real time series data for Diffusion-TS ($\lambda=0.1$).}
    \label{fig:tsne}
\end{figure}

\begin{figure}[h!]
    \centering
    \includegraphics[width=1.0\linewidth]{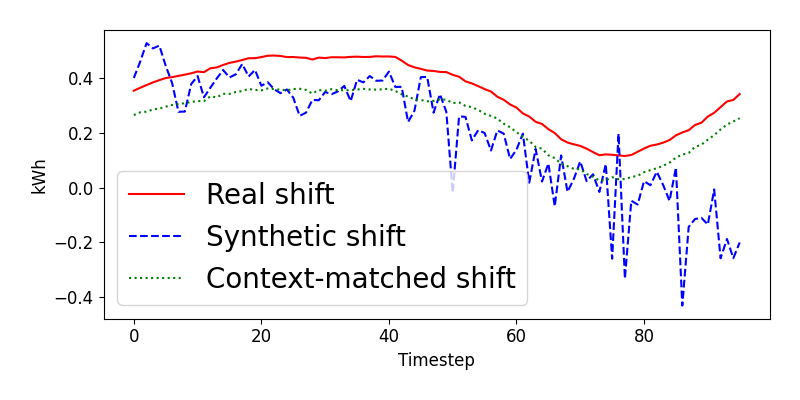}
    \vspace{-0.3cm}
    \includegraphics[width=1.0\linewidth]{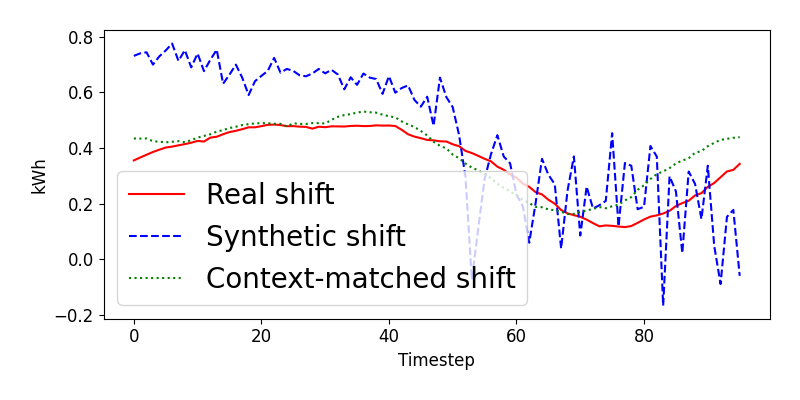}
    \vspace{-0.3cm}
    \includegraphics[width=1.0\linewidth]{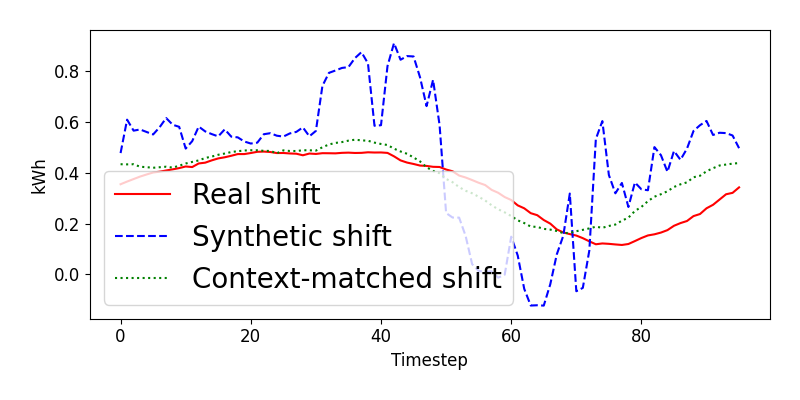}
    \vspace{-0.3cm}
    \includegraphics[width=1.0\linewidth]{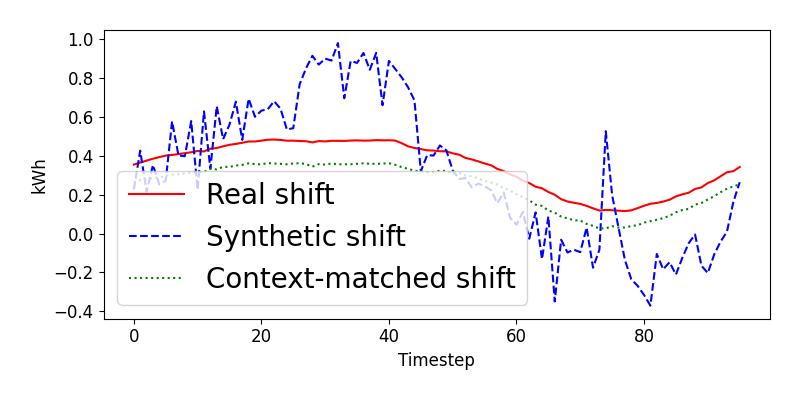}
    \caption{Shift of time series values from no pv households to pv households for CENTS with ACGAN (Top 2) and CENTS with Diffusion-TS (Bottom 2).}
    \label{fig:diffusion-shift-appendix}
\end{figure}

\begin{figure*}[h!]
    \centering
    \begin{subfigure}[t]{0.49\linewidth}
        \centering
        \includegraphics[width=\linewidth]{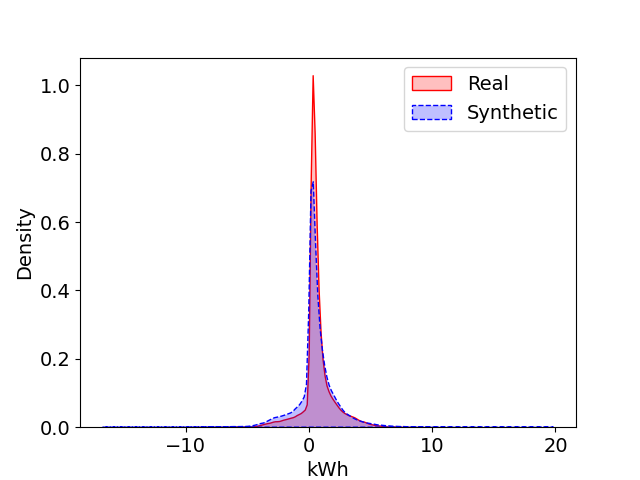}
        \caption{Kernel density estimate for synthetic and real time series for ACGAN ($\lambda=0.1$).}
        \label{fig:kde_acgan}
    \end{subfigure}
    \hfill
    \begin{subfigure}[t]{0.49\linewidth}
        \centering
        \includegraphics[width=\linewidth]{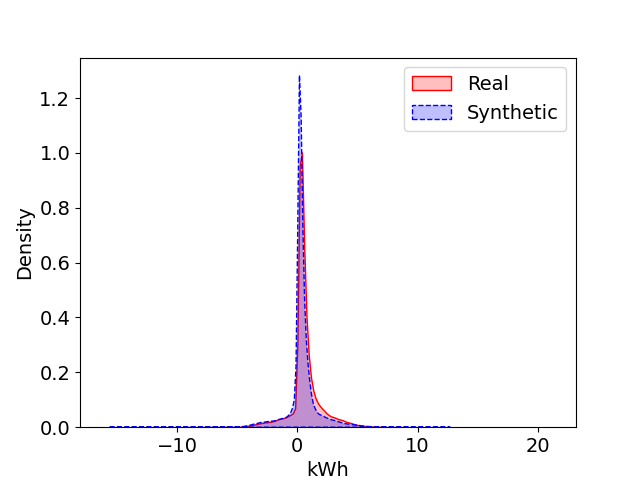}
        \caption{Kernel density estimate for synthetic and real time series for Diffusion-TS ($\lambda=0.1$).}
        \label{fig:kde_diffusion}
    \end{subfigure}
    \caption{Comparison of Kernel Density Estimate (KDE) plots for synthetic kWh values.}
    \label{fig:kde_comparison}
\end{figure*}

\begin{figure*}[h!]
    \centering
        \begin{subfigure}[t]{0.7\linewidth}
        \centering
        \includegraphics[width=\linewidth]{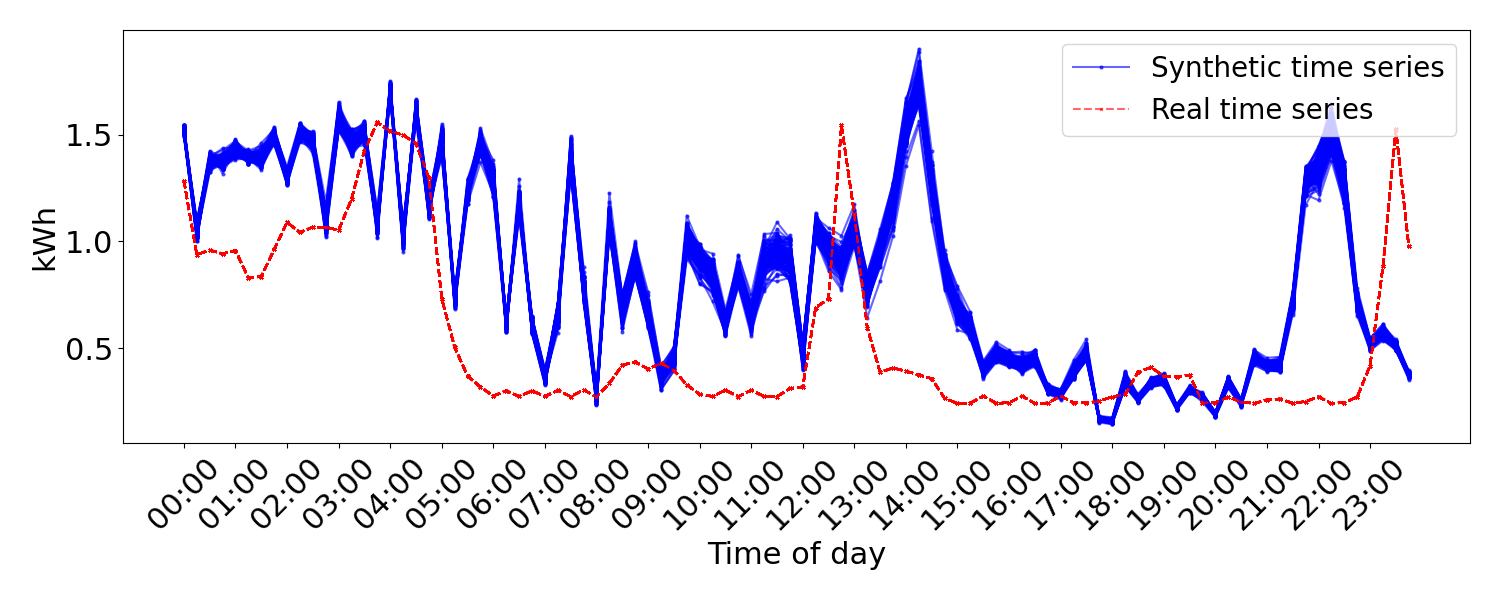}
        \label{fig:acgan_comp_1a}
    \end{subfigure}
    \hfill
    \begin{subfigure}[t]{0.7\linewidth}
        \centering
        \includegraphics[width=\linewidth]{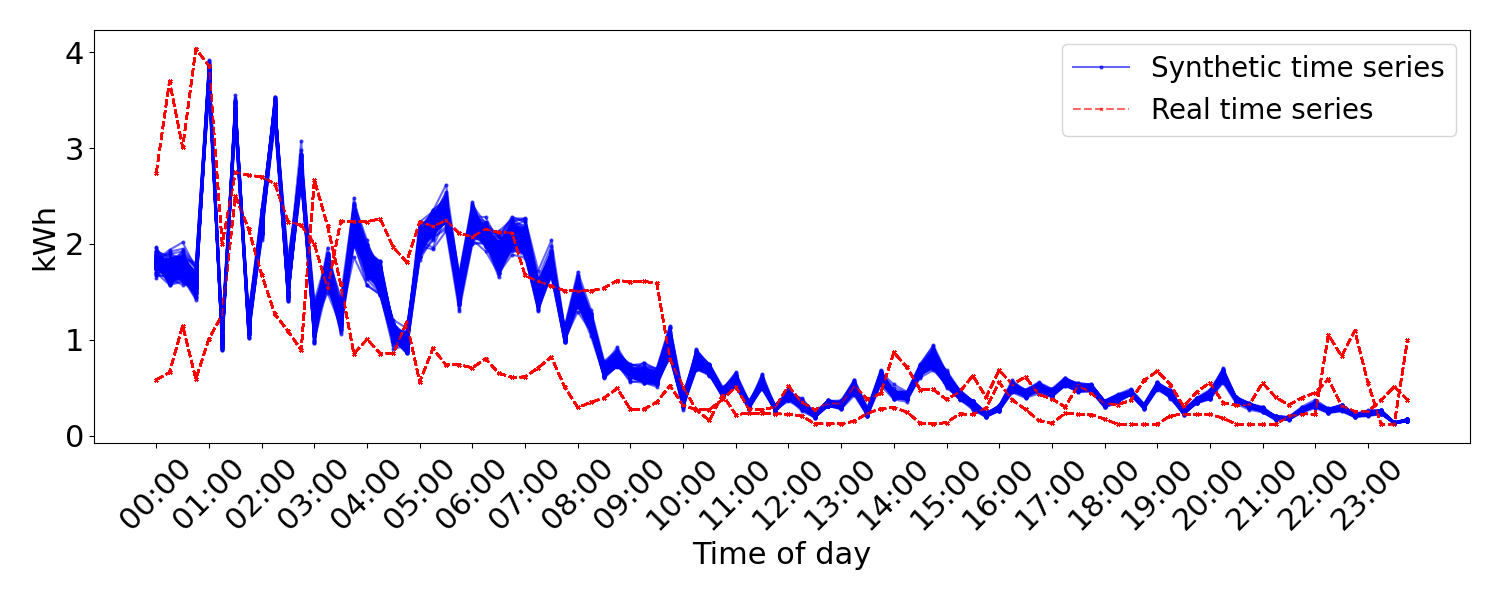}
        \label{fig:acgan_comp_2a}
    \end{subfigure}
    \begin{subfigure}[t]{0.7\linewidth}
        \centering
        \includegraphics[width=\linewidth]{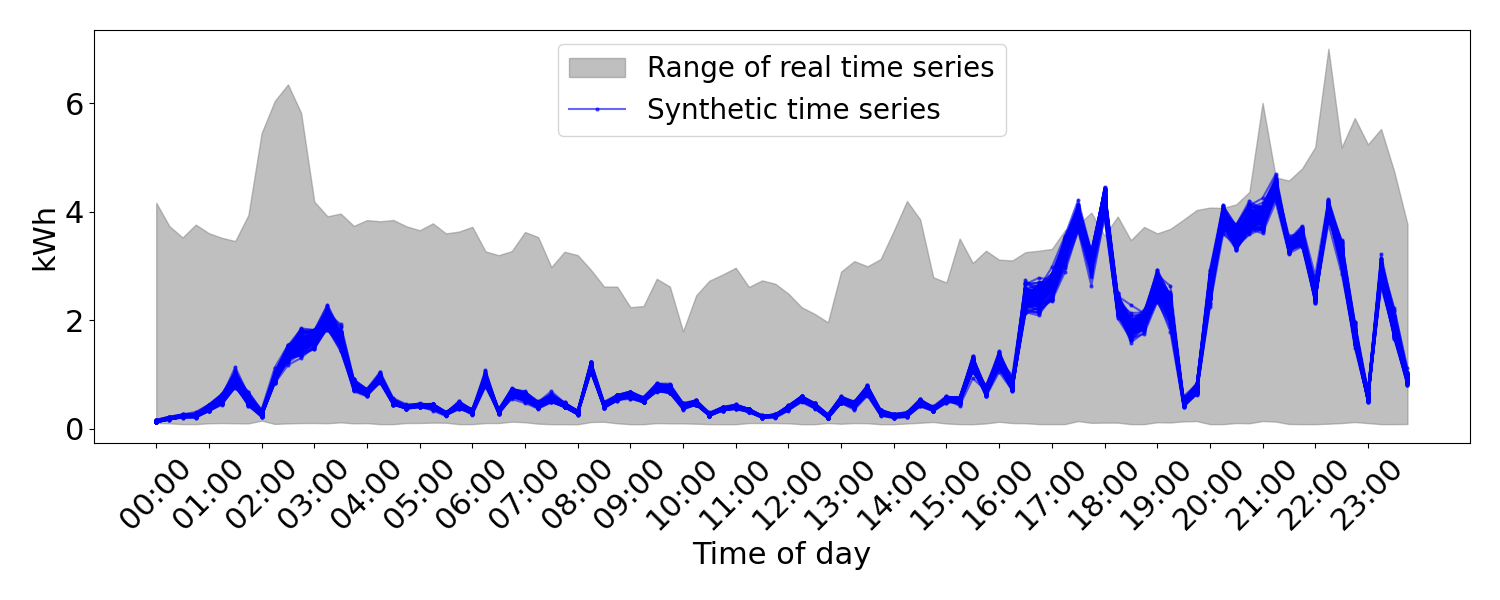}
        \label{fig:acgan_range_1a}
    \end{subfigure}
    \hfill
    \begin{subfigure}[t]{0.7\linewidth}
        \centering
        \includegraphics[width=\linewidth]{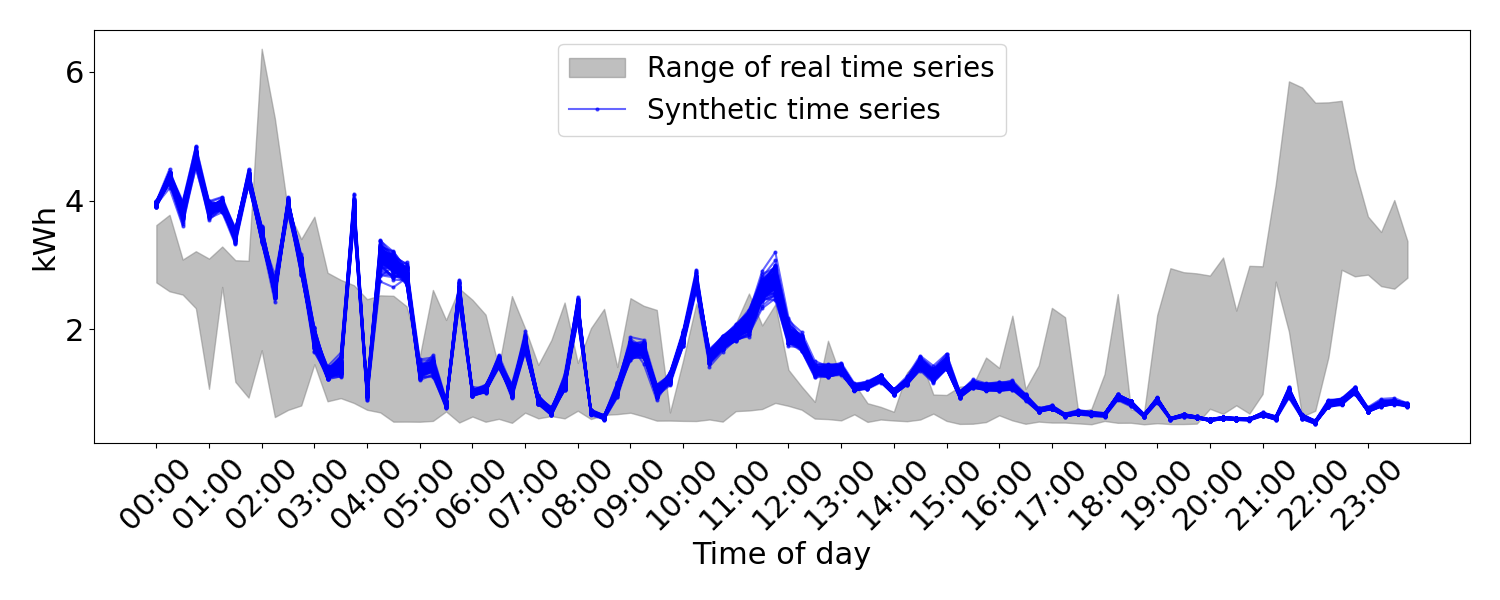}
        \label{fig:acgan_range_2a}
    \end{subfigure}
    \caption{Range of real data vs 100 synthetically generated time series for the same context using CENTS with ACGAN. Synthetic time series are highly realistic but show limited variance and mode coverage.}
    \label{fig:acgan_app}
\end{figure*}

\begin{figure*}[h!]
    \centering
        \begin{subfigure}[t]{0.7\linewidth}
        \centering
        \includegraphics[width=\linewidth]{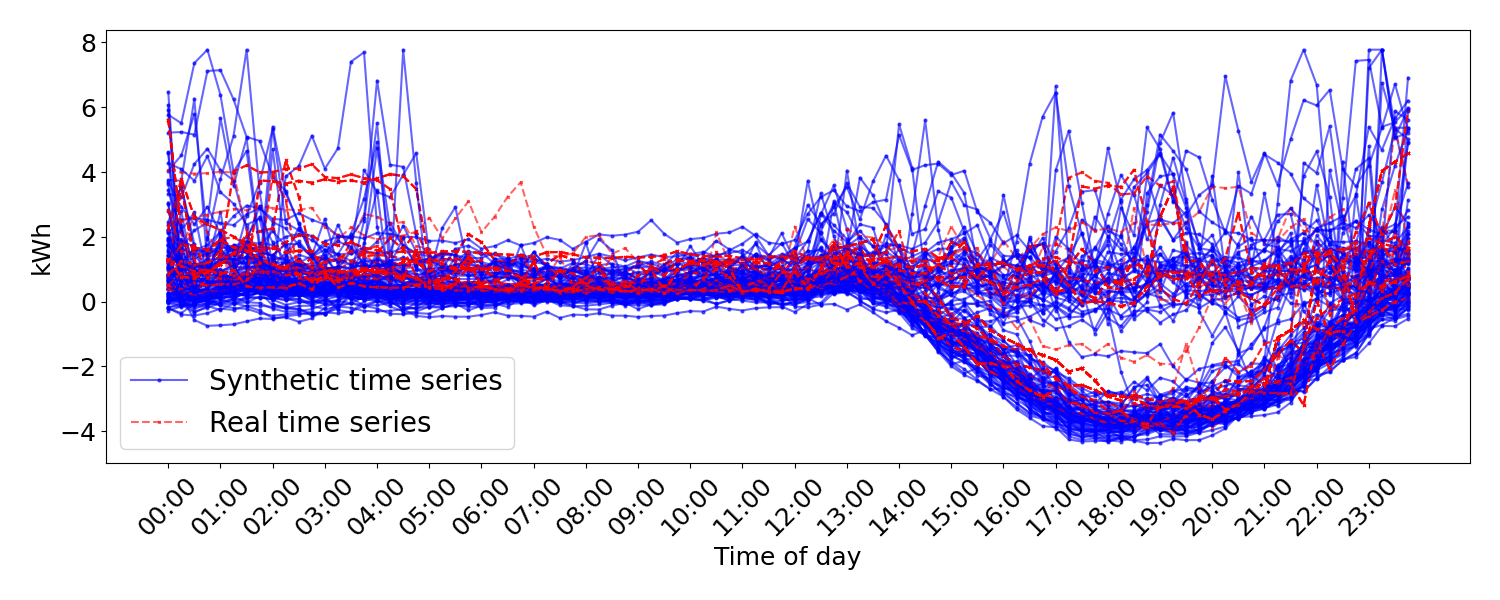}
        \label{fig:diff_comp_app1}
    \end{subfigure}
    \hfill
    \begin{subfigure}[t]{0.7\linewidth}
        \centering
        \includegraphics[width=\linewidth]{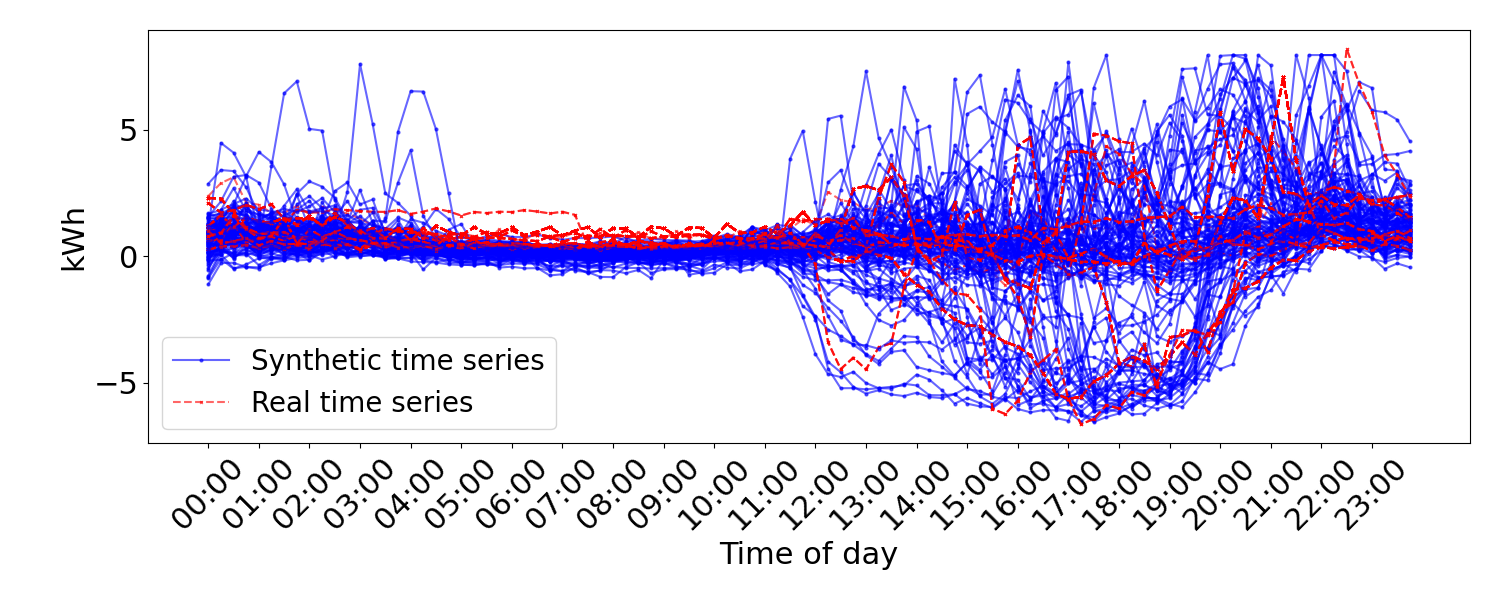}
        \label{fig:diff_comp_app2}
    \end{subfigure}
    \begin{subfigure}[t]{0.7\linewidth}
        \centering
        \includegraphics[width=\linewidth]{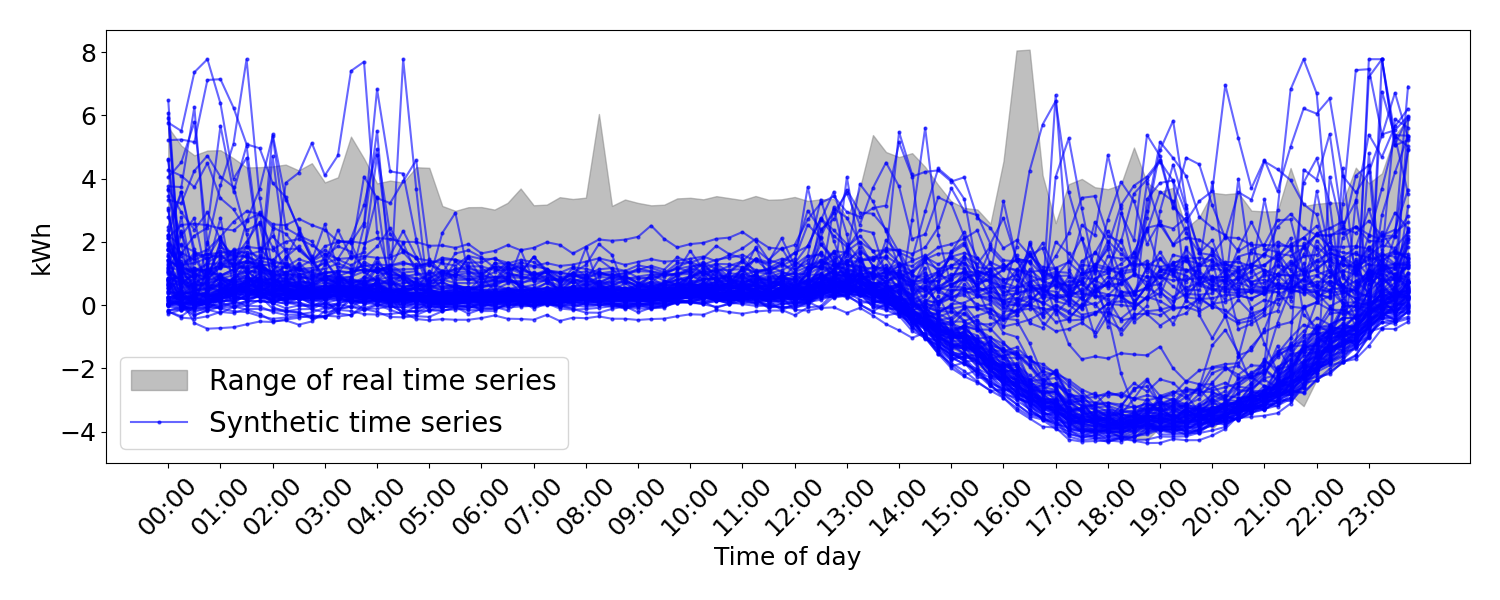}
        \label{fig:diff_range_app1}
    \end{subfigure}
    \hfill
    \begin{subfigure}[t]{0.7\linewidth}
        \centering
        \includegraphics[width=\linewidth]{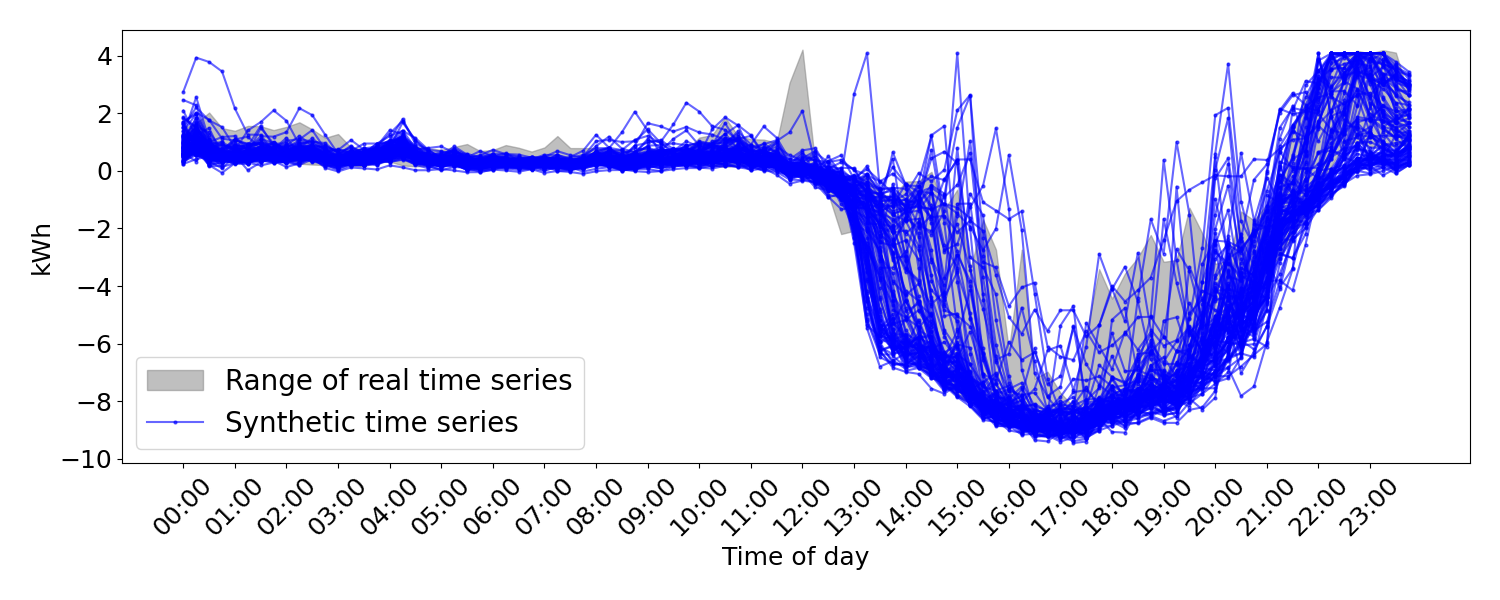}
        \label{fig:diff_range_app2}
    \end{subfigure}
    \caption{Range of real data vs 100 synthetically generated time series for the same context using CENTS with Diffusion-TS. Synthetic time series capture the multimodal distribution and are highly realistic.}
    \label{fig:diffusion_app}
\end{figure*}

\end{document}